\documentclass[11pt,a4paper]{article}
\usepackage[hyperref]{acl2020}
\usepackage{times}
\usepackage{latexsym}
\usepackage{amsmath}
\usepackage{amssymb}
\usepackage{booktabs}
\usepackage{graphicx}
\usepackage{url}
\usepackage{xcolor}

\DeclareMathOperator*{\argmin}{arg\,min}
\DeclareMathOperator*{\FC}{FC}
\DeclareMathOperator*{\VGG}{VGG-16}
\DeclareMathOperator*{\ReLU}{ReLU}
\newcommand*{\LNL}{\ensuremath{\mathcal{L}_{\text{NL}}}}
\newcommand*{\LCLS}{\ensuremath{\mathcal{L}_{\text{CLS}}}}
\newcommand*{\LAMBDANL}{\ensuremath{\lambda_{\text{NL}}}}

\usepackage{microtype}

\aclfinalcopy % Uncomment this line for the final submission
\newif\ifcomments

\title{Shaping Visual Representations with Language \\ for Few-Shot Classification}

\author{Jesse Mu{\normalfont \textsuperscript{1}},$\;$ Percy Liang{\normalfont \textsuperscript{1}},$\;$ Noah D.\ Goodman{\normalfont \textsuperscript{1,2}} \\
  Departments of \textsuperscript{1}Computer Science and \textsuperscript{2}Psychology \\
  Stanford University \\
  {\tt \{muj,ngoodman\}@stanford.edu, pliang@cs.stanford.edu} \\
}

\date{}

\begin{document}
\maketitle
\begin{abstract}
By describing the features and abstractions of our world, language is a crucial tool for human learning and a promising source of supervision for machine learning models.
  We use language to improve few-shot visual classification in the underexplored scenario where natural language task descriptions are available during training, but unavailable for novel tasks at test time.
Existing models for this setting sample new descriptions at test time and use those to classify images.
Instead, we propose \emph{language-shaped learning} (LSL), an end-to-end model that regularizes visual representations to predict language.
LSL is conceptually simpler, more data efficient, and outperforms baselines in two challenging few-shot domains.
\end{abstract}

\section{Introduction}

Humans are powerful and efficient learners partially due to the ability to \emph{learn from language} \cite{chopra2019ratchet,tomasello1999cultural}. For instance, we can learn about \emph{robins} not by seeing thousands of examples, but by being told that \emph{a robin is a bird with a red belly and brown feathers}. This language further shapes the way we view the world, constraining our hypotheses for new concepts: given a new bird (e.g.\ \emph{seagulls}), even without language we know that features like belly and feather color are relevant \citep{goodman1955fact}.

\begin{figure}[t]
  \centering
  \includegraphics[width=\linewidth]{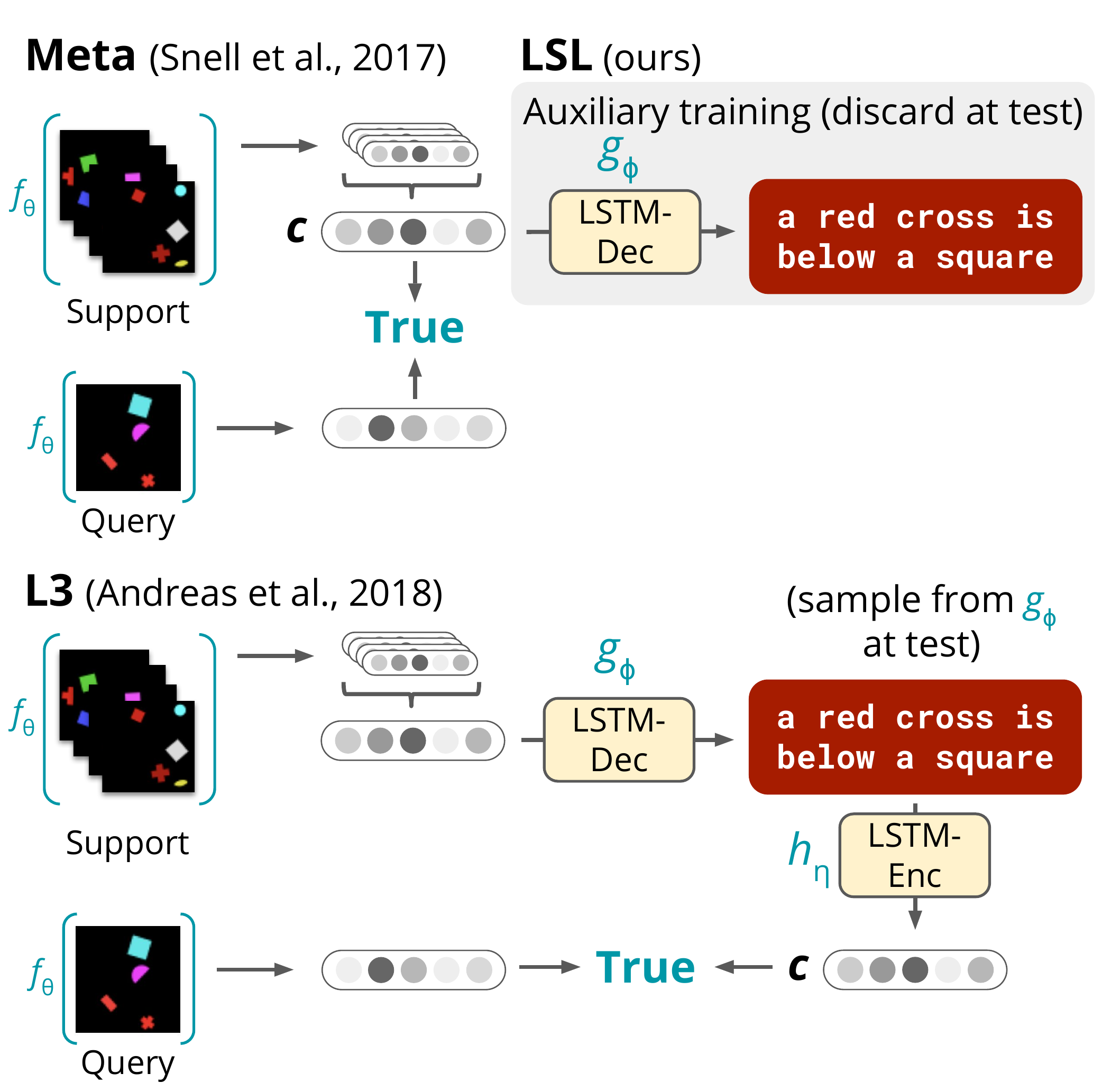}
  \caption{We propose few-shot classification models whose learned representations are constrained to predict natural language task descriptions during training, in contrast to models which explicitly use language as a bottleneck for classification \citep{andreas2018learning}.}
  \label{fig:overview}
\end{figure}

In this paper, we guide visual representation learning with language, studying the setting where \emph{no language is available at test time}, since rich linguistic supervision is often unavailable for new concepts encountered in the wild.
How can one best use language in this setting?
One option is to just regularize, training representations to predict language descriptions.
Another is to exploit the compositional nature of language directly by using it as a bottleneck in a discrete latent variable model.
% as a bottleneck during inference.
For example, the recent \emph{Learning with Latent Language} \cite[L3;][]{andreas2018learning} model does both: during training, language is used to classify images; at test time, with no language, descriptions are sampled from a decoder conditioned on the language-shaped image embeddings.

Whether the bottleneck or regularization most benefits models like L3 is unclear.
We disentangle these effects and propose \emph{language-shaped learning} (LSL), an end-to-end model that uses visual representations shaped by language (Figure~\ref{fig:overview}), thus avoiding the bottleneck.
We find that discrete bottlenecks can hurt performance, especially with limited language data; in contrast, LSL is architecturally simpler, faster, uses language more efficiently, and outperforms L3 and baselines across two few-shot transfer tasks.

\section{Related Work}

Language has been shown to assist visual classification in various settings, including traditional visual classification with no transfer \citep{he2017fine} and with language available at test time in the form of class labels or descriptions for zero- \cite{frome2013devise,socher2013zero} or few-shot \cite{xing2019adaptive} learning.
Unlike past work, we have no language at test time and test tasks differ from training tasks, so language from training cannot be used as additional class information \citep[cf.][]{he2017fine} or weak supervision for labeling additional in-domain data \citep[cf.][]{hancock2018training}.
Our setting can be viewed as an instance of \emph{learning using privileged information} \citep[LUPI;][]{vapnik2009new}, where richer supervision augments a model only during training.

In this framework, learning with attributes and other domain-specific rationales has been tackled extensively \cite{zaidan2007using,donahue2011annotator,tokmakov2019learning}; language less so. 
\citet{gordo2017beyond} use METEOR scores between captions as a similarity measure for specializing embeddings for image retrieval, but do not directly ground language explanations.
\citet{srivastava2017joint} explore a supervision setting similar to ours, except in simple text and symbolic domains where descriptions can be easily converted to executable logical forms via semantic parsing.

Another line of work studies the generation of natural language explanations for interpretability across language \citep[e.g.\ entailment;][]{camburu2018snli} and vision \citep{hendricks2016generating,hendricks2018grounding} tasks, but here we examine whether predicting language can actually improve task performance; similar ideas have been explored in text \cite{rajani2019explain} and reinforcement learning \cite{bahdanau2019learning,goyal2019using} domains.

\section{Language-shaped learning}

We are interested in settings where language explanations can help learn representations that generalize more efficiently across tasks, especially when training data for each task is scarce and there are many spurious hypotheses consistent with the input. Thus, we study the few-shot (meta-)learning setting, where a model must learn from a set of train tasks, each with limited data, and then generalize to unseen tasks in the same domain.

Specifically, in $N$-way, $K$-shot learning, a task $t$ consists of $N$ \emph{support} classes $\{ \mathcal{S}^{(t)}_1, \dots, \mathcal{S}^{(t)}_N \}$ with $K$ examples each: $\mathcal{S}^{(t)}_n = \{ \mathbf{x}^{(t)}_{n, 1}, \dots, \mathbf{x}^{(t)}_{n, K} \}$.
Each task has $M$ \emph{query} examples 
$\mathcal{Q}^{(t)} = \{(\mathbf{x}^{(t)}_1, y^{(t)}_1), \dots, (\mathbf{x}^{(t)}_M, y^{(t)}_M)\}$.
Given the $m$-th query example $\mathbf{x}^{(t)}_m$ as input, the goal is to predict its class $y^{(t)}_m \in \{1, \dots, N\}$. After learning from a set of tasks $\mathcal{T}_{\text{train}}$, a model is evaluated on unseen tasks $\mathcal{T}_{\text{test}}$.

While the language approach we propose is applicable to nearly any meta-learning framework, we use prototype networks \citep{snell2017prototypical}, which have a simple but powerful inductive bias for few-shot learning.
Prototype networks learn an embedding function
$f_\theta$ for examples;
the embeddings of the support examples of a class $n$ are averaged to form a class \emph{prototype}  (omitting task $^{(t)}$ for clarity):
\begin{align}
  \mathbf{c}_n = \frac{1}{K} \sum_{k = 1}^K f_\theta(\mathbf{x}_{n, k}).
\end{align}
Given a query example $(\mathbf{x}_m, y_m)$, we predict class $n$ with probability proportional to some similarity function $s$ between $\mathbf{c}_n$ and $f_\theta(\mathbf{x}_m)$:
\begin{align}
  p_\theta(\hat{y}_m = n \mid \mathbf{x}_m) \propto \exp \left(s\left( \mathbf{c}_n, f_\theta \left( \mathbf{x}_m \right) \right) \right).
\end{align} 
$f_\theta$ is then trained to minimize the \emph{classification loss}
\begin{align}
\LCLS(\theta) &= -\sum_{m = 1}^M \log p_\theta\left( \hat{y}_m = y_m \mid \mathbf{x}_m \right).
\end{align}

\subsection{Shaping with language} Now assume that during training we have for each class $\mathcal{S}_n$ a set of $J_n$ associated natural language descriptions $\mathcal{W}_n = \{ \mathbf{w}_1, \dots, \mathbf{w}_{J_n} \}$.
Each $\mathbf{w}_j$ should explain the relevant features of $\mathcal{S}_n$ and need not be associated with individual examples.\footnote{If we have language associated with individual examples, we can regularize at the instance-level, essentially learning an image captioner. We did not observe major gains with instance-level supervision (vs class-level) in the tasks explored here, in which case class-level language is preferable, since it is much easier to obtain. There are likely tasks where instance-level supervision is superior, which we leave for future work.}
In Figure~\ref{fig:overview}, we have one description $\mathbf{w}_1 = ( \texttt{A}, \texttt{red}, \dots, \texttt{square})$.

Our approach is simple: we encourage $f_\theta$ to learn prototypes that can also decode the class language descriptions. Let $\tilde{\mathbf{c}}_n$ be the prototype formed by averaging the support \emph{and} query examples of class $n$. Then define a language model $g_\phi$ (e.g., a recurrent neural network), which conditioned on $\tilde{\mathbf{c}}_n$ provides a probability distribution over descriptions $g_\phi(\hat{\mathbf{w}}_{j} \mid \tilde{\mathbf{c}}_n)$
with a corresponding \emph{natural language loss}:
\begin{align}
  &\LNL(\theta, \phi) = - \sum_{n=1}^{N} \sum_{j=1}^{J_n} \log g_\phi (\mathbf{w}_{j} \mid \tilde{\mathbf{c}}_n),
\end{align}
i.e.\ the total negative log-likelihood of the class descriptions across all classes in the task. Since $\LNL$ depends on parameters $\theta$ through the prototype $\tilde{\mathbf{c}}_n$, this objective should encourage our model to better represent the features expressed in language.

Now we jointly minimize both losses:
\begin{align}
    \argmin_{\theta, \phi} \left[ \LCLS(\theta) + \LAMBDANL \LNL(\theta, \phi) \right],
\end{align}
where the hyperparameter $\LAMBDANL$ controls the weight of the natural language loss. At test time, we simply discard $g_\phi$ and use $f_\theta$ to classify. We call our approach \emph{language-shaped learning} (LSL; Figure~\ref{fig:overview}).

\subsection{Relation to L3}
L3 \citep{andreas2018learning} has the same basic components of LSL, but instead defines the concepts 
$\mathbf{c}_n$ to be embeddings of the language descriptions themselves, generated by an additional recurrent neural network (RNN) encoder $h_\eta$: $\mathbf{c}_n = h_\eta(\mathbf{w}_n)$. During training, the ground-truth description is used for classification, while $g_\phi$ is trained to produce the description; at test time, L3 \emph{samples} candidate descriptions $\hat{\mathbf{w}}_n$ from $g_\phi$, keeping the description most similar to the images in the support set according to the similarity function $s$ (Figure~\ref{fig:overview}).

Compared to L3, LSL is simpler since it (1) does not require the additional embedding module $h_\eta$ and (2) does not need the test-time language sampling procedure.\footnote{LSL is similar to the ``Meta+Joint'' model of \citet{andreas2018learning}, which did not improve over baseline. However, they used separate encoders for the support and query examples, with only the support encoder trained to predict language, resulting in overfitting of the query encoder.} This also makes LSL much faster to run than L3 in practice: without the language machinery, LSL is up to 50x faster during inference in our experiments.

\section{Experiments}

Here we describe our two tasks and models. For each task, we evaluate LSL, L3, and a prototype network baseline trained without language (Meta; Figure~\ref{fig:overview}). For full details, see Appendix~\ref{app:details}.

\begin{figure}[t]
  \centering
  \includegraphics[width=0.8\linewidth]{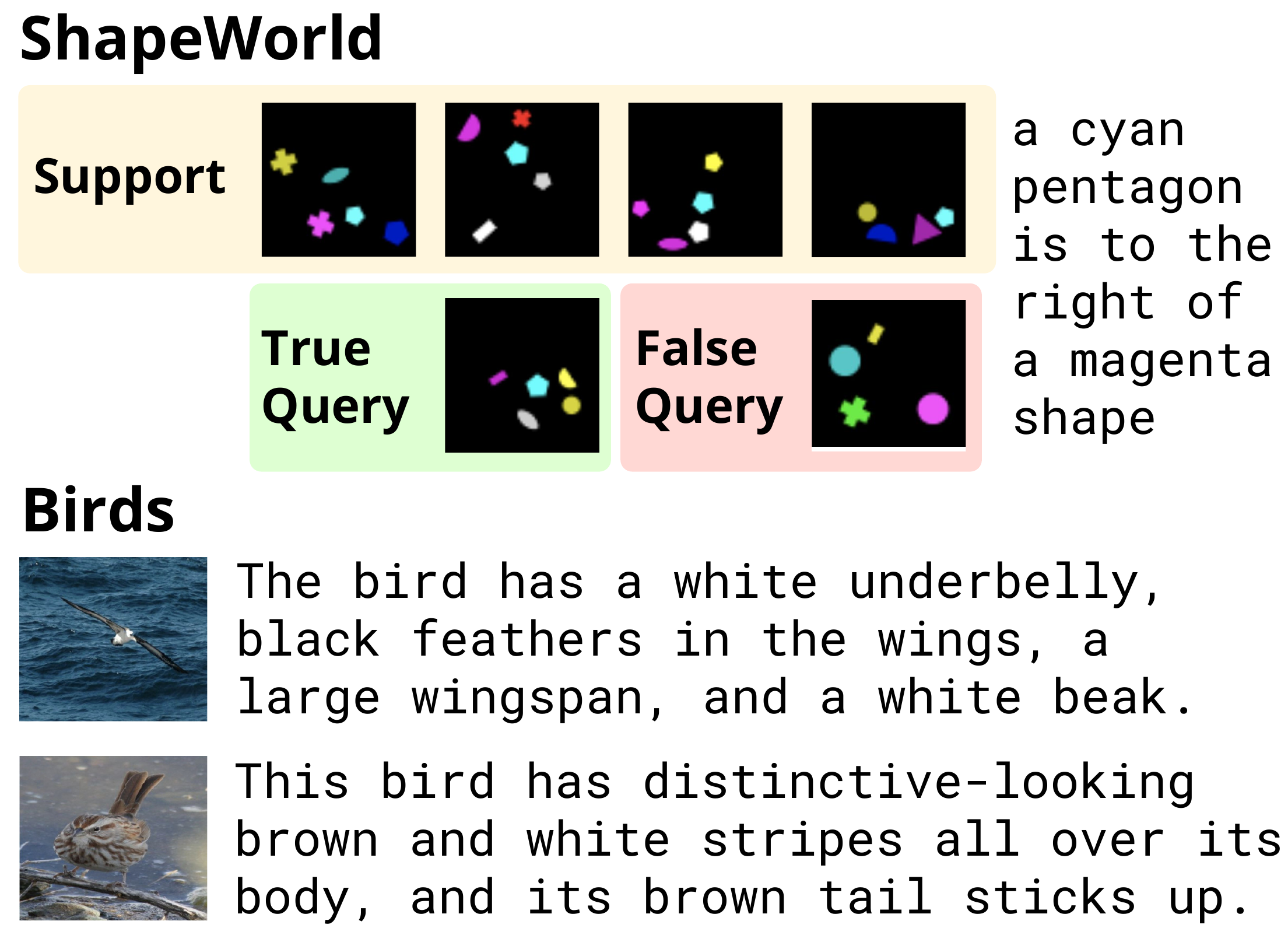}
  \caption{Example language and query examples for ShapeWorld and Birds.}
  \label{fig:example_language}
\end{figure}

\paragraph{ShapeWorld.}
First, we use the ShapeWorld \citep{kuhnle2017shapeworld} dataset used by \citet{andreas2018learning}, which consists of 9000 training, 1000 validation, and 4000 test tasks (Figure~\ref{fig:example_language}).\footnote{This is a larger version with 4x as many test tasks for more stable confidence intervals (see Appendix~\ref{app:details}).} Each task contains a single support set of $K = 4$ images representing a visual concept with an associated (artificial) English language description, generated with a minimal recursion semantics representation of the concept \cite{copestake2016resources}. Each concept is a spatial relation between two objects, each object optionally qualified by color and/or shape, with 2-3 distractor shapes present. The task is to predict whether a query image $\mathbf{x}$ belongs to the concept.

For ease of comparison, we report results with models identical to \citet{andreas2018learning}, where $f_\theta$ is the final convolutional layer of a fixed ImageNet-pretrained VGG-16 \cite{simonyan2014very} fed through two fully-connected layers:
\begin{align}
    f_\theta(\mathbf{x}) = \FC(\ReLU(\FC(\VGG(\mathbf{x})))).
\end{align}

However, because fixed ImageNet representations may not be the most appropriate choice for artificial data, we also run experiments with convolutional networks trained from scratch: either the 4-layer convolutional backbone used in much of the few-shot literature \cite{chen2019closer}, as used in the Birds experiments we describe next, or a deeper ResNet-18 \cite{he2016deep}.

This is a special binary case of the few-shot learning framework, with a single positive support class $\mathcal{S}$ and prototype $\mathbf{c}$.
Thus, we define the similarity function to be the sigmoid function $s(a, b) = \sigma(a \cdot b)$ and the positive prediction $P(\hat{y} = 1 \mid \mathbf{x}) = s\left(f_\theta(\mathbf{x}), \mathbf{c}\right)$. 
$g_\phi$ is a 512-dimensional gated recurrent unit (GRU) RNN \cite{cho2014learning} trained with teacher forcing. Through a grid search on the validation set, we set $\LAMBDANL = 20$.

\paragraph{Birds.}
To see if LSL can scale to more realistic scenarios, we use the Caltech-UCSD Birds dataset \cite{wah2011caltech}, which contains 200 bird species, each with 40--60 images, split into 100 train, 50 validation, and 50 test classes. During training, tasks are sampled dynamically by selecting $N$ classes from the 100 train classes. $K$ support and 16 query examples are then sampled from each class (similarly for val and test).
For language, we use the descriptions collected by \citet{reed2016learning}, where AMT crowdworkers were asked to describe individual images of birds in detail, without reference to the species (Figure~\ref{fig:example_language}).

While 10 English descriptions per image are available, we assume a more realistic scenario where we have \emph{much less} language available only at the class level: removing associations between images and their descriptions, we aggregate $D$ descriptions for each class, and for each $K$-shot training task we sample $K$ descriptions from each class $n$ to use as descriptions $\mathcal{W}_n$.
This makes learning especially challenging for LSL due to noise from captions that describe features only applicable to individual images.
Despite this, we found improvements with as few as $D = 20$ descriptions per class, which we report as our main results, but also vary $D$ to see how efficiently the models use language.

We evaluate on the $N = 5$-way, $K = 1$-shot setting, and as $f_\theta$ use the 4-layer convolutional backbone proposed by \citet{chen2019closer}. Here we use a learned bilinear similarity function, $s(a, b) = a^{\top}\mathbf{W}b$, where $\mathbf{W}$ is learned jointly with the model. $g_\phi$ is a 200-dimensional GRU, and with another grid search we set $\LAMBDANL = 5$.

\section{Results}

\begin{figure}[t]
  \centering
  \includegraphics[width=0.95\linewidth]{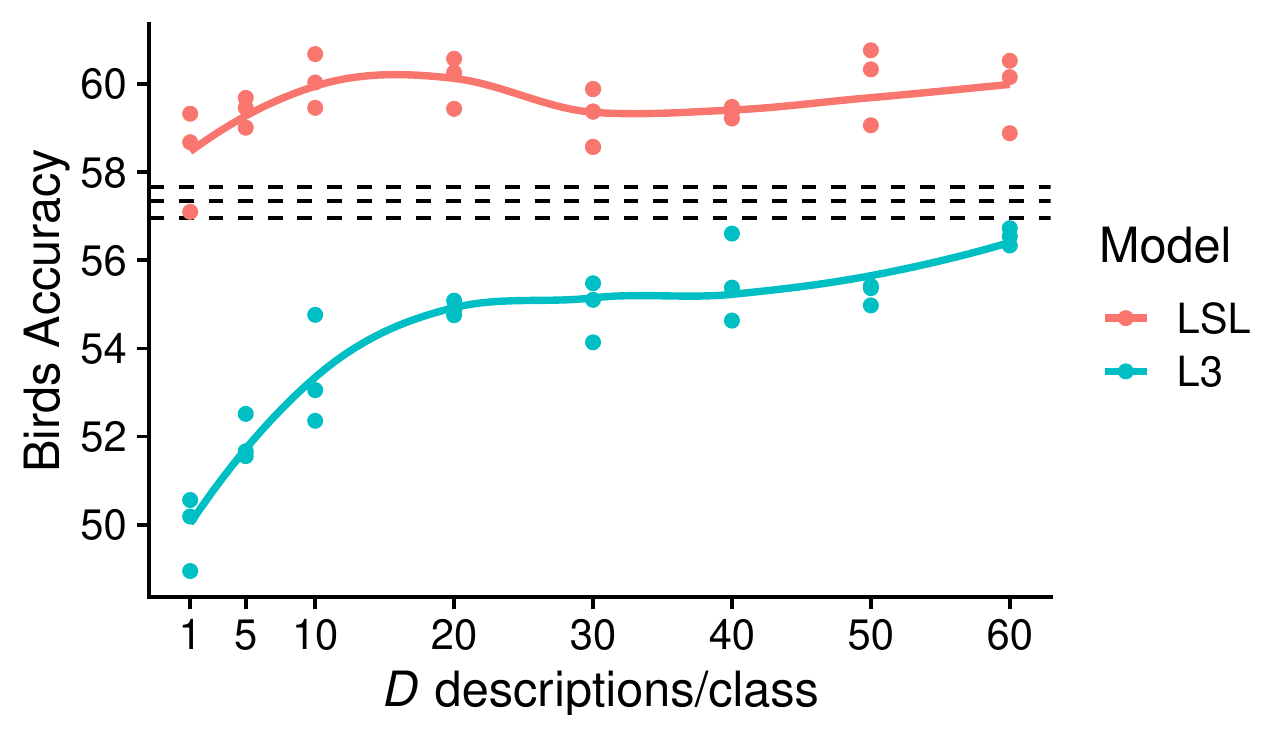}
  \caption{Varying the descriptions per class, $D$, for Birds. Each dot is a separate independently trained model. The dashed lines represent independently trained baselines (Meta).}
  \label{fig:language_amount}
\end{figure}

Results are in Table~\ref{tab:main_results}. For ShapeWorld, LSL outperforms the meta-learning baseline (Meta) by 6.7\%, and does at least as well as L3; Table~\ref{tab:conv_results} shows similar trends when $f_\theta$ is trained from scratch.
For Birds, LSL has a smaller but still significant 3.3\% increase over Meta, while L3 drops below baseline.
Furthermore, LSL uses language more efficiently: Figure~\ref{fig:language_amount} shows Birds performance as the captions per class $D$ increases from 1 (100 total) to 60 (6000 total). LSL benefits from a remarkably small number of captions, with limited gains past 20; in contrast, L3 requires much more language to even approach baseline performance.

\begin{figure}[t]
  \centering
  \includegraphics[width=\linewidth]{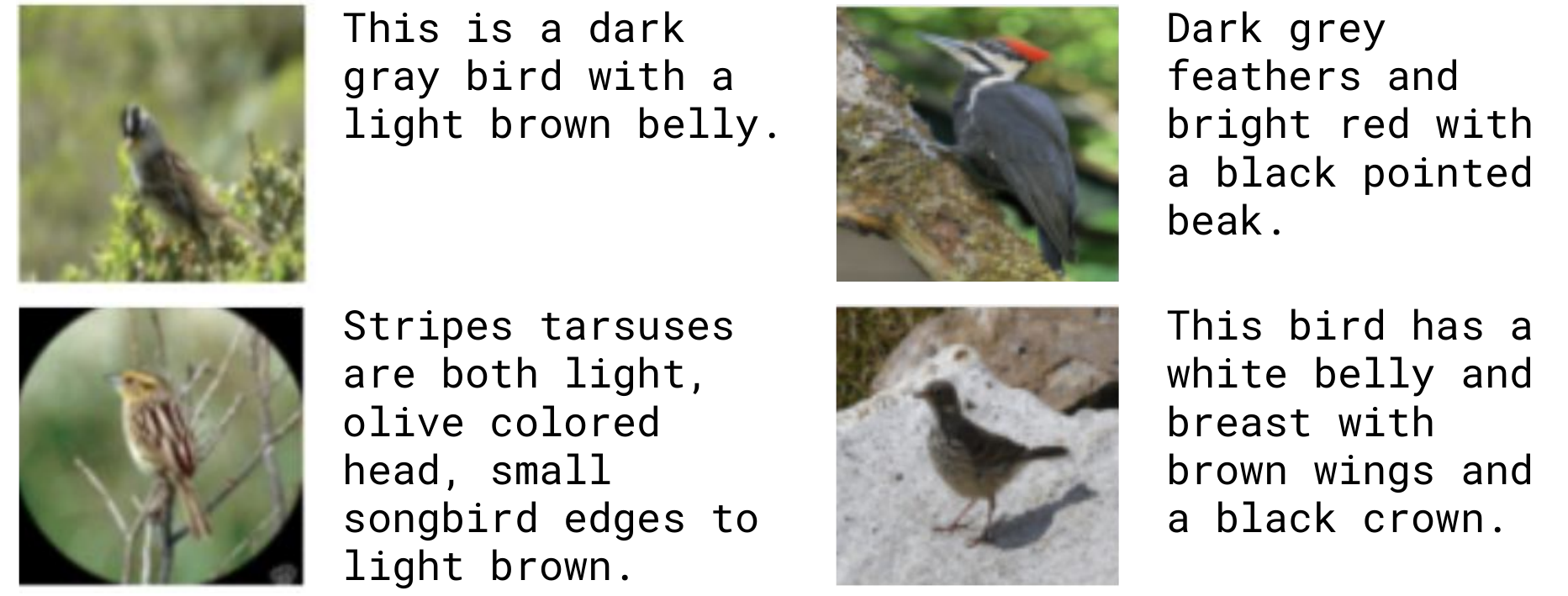}
  \caption{Examples of language generated by the L3 decoder $g_\phi$ for Birds validation images. Since the LSL decoder is identically parameterized, it generates similar language.}
  \label{fig:language_examples}
\end{figure}

In the low-data regime, L3's lower performance is unsurprising, since it must generate language at test time, which is difficult with so little data.
Example output from the L3 decoder in Figure~\ref{fig:language_examples} highlights this fact: the language looks reasonable in some cases, but in others has factual errors (\emph{dark gray bird; black pointed beak}) and fluency issues.

These results suggest that any benefit of L3 is likely due to the regularizing effect that language has on its embedding model $f_\theta$, which has been trained to predict language for test-time inference; in fact, the discrete bottleneck actually hurts in some settings.
By using only the regularized visual representations and not relying exclusively on the generated language, LSL is the simpler, more efficient, and overall superior model.

\begin{table}
\centering
\caption{Test accuracies ($\pm$ 95\% CI) across 1000 (ShapeWorld) and 600 (Birds) tasks.}
\label{tab:main_results}
\begin{tabular}{lrr}
\toprule
    & ShapeWorld & Birds $(D = 20)$ \\
\midrule
  Meta & 60.59 $\pm$ 1.07 & 57.97 $\pm$ 0.96 \\
  L3 & 66.60 $\pm$ 1.18 & 53.96 $\pm$ 1.06 \\
  LSL & \textbf{67.29 $\pm$ 1.03} & \textbf{61.24 $\pm$ 0.96} \\
\bottomrule
\end{tabular}
\end{table}

\begin{table}
\centering
\caption{ShapeWorld performance with different $f_\theta$ architectures trained from scratch.}
\label{tab:conv_results}
\begin{tabular}{lrr}
\toprule
 $f_\theta$   & Conv4 & ResNet-18 \\
\midrule
  Meta & 50.91 $\pm$ 1.10 & 58.73 $\pm$ 1.08 \\
  L3 & 62.28 $\pm$ 1.09 & 67.90 $\pm$ 1.07 \\
  LSL & \textbf{63.25 $\pm$ 1.06} & \textbf{68.76 $\pm$ 1.02} \\
\bottomrule
\end{tabular}
\end{table}

\begin{figure}[t]
  \centering
  \includegraphics[width=0.9\linewidth]{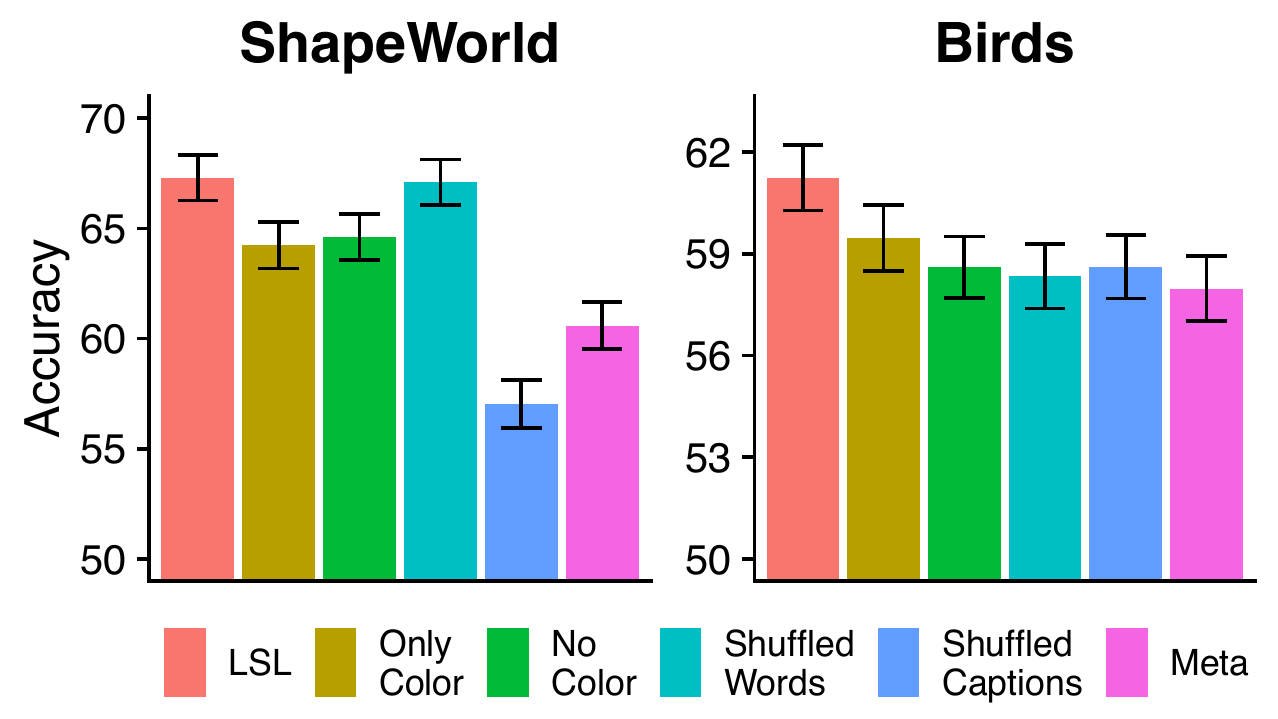}
  \caption{Language ablations. Error bars are 95\% CIs.}
  \label{fig:lang_variants}
\end{figure}

\subsection{Language ablation}
\begin{figure}[t]
  \centering
  \includegraphics[width=\linewidth]{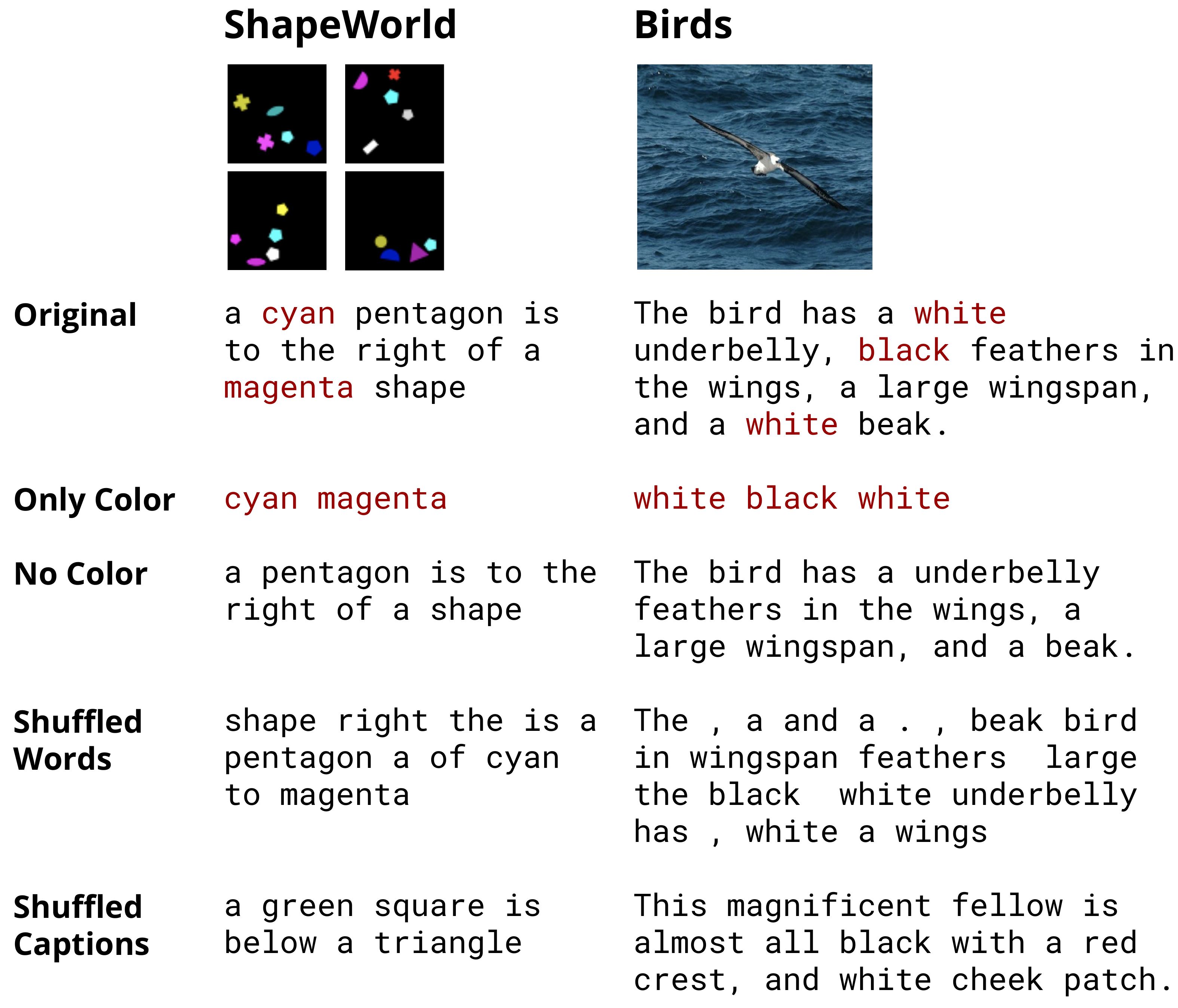}
  \caption{Examples of ablated language supervision for the Birds and ShapeWorld tasks.}
  \label{fig:lang_ablation_examples}
\end{figure}

To identify which aspects of language are most helpful, in Figure~\ref{fig:lang_variants} we examine LSL performance under ablated language supervision: (1) keeping only a list of common color words, (2) filtering out color words, (3) shuffling the words in each caption, and (4) shuffling the captions across tasks (see Figure~\ref{fig:lang_ablation_examples} for examples).

We find that while the benefits of color/no-color language varies across tasks, neither component provides the benefit of complete language, demonstrating that LSL leverages both colors and other attributes (e.g.\ size, shape) described in language.
Word order is important for Birds but surprisingly unimportant for ShapeWorld, suggesting that even with decoupled colors and shapes, the model can often infer the correct relation from the shapes that consistently appear in the examples.
Finally, when captions are shuffled across tasks, LSL for Birds does no worse than Meta, while ShapeWorld suffers, suggesting that language is more important for ShapeWorld than for the fine-grained, attribute-based Birds task.

\section{Discussion}

We presented LSL, a few-shot visual recognition model that is regularized with language descriptions during training. LSL outperforms baselines across two tasks and uses language supervision more efficiently than L3.
We find that if a model is trained to expose the features and abstractions in language, a linguistic bottleneck on top of these language-shaped representations is unnecessary, at least for the kinds of visual tasks explored here.

The line between language and sufficiently rich attributes and rationales is blurry, and recent work \cite{tokmakov2019learning} suggests that similar performance gains can likely be observed by regularizing with attributes.
However, unlike attributes, language is (1) a more natural medium for annotators, (2) does not require preconceived restrictions on the kinds of features relevant to the task, and (3) is abundant in unsupervised forms.
This makes shaping representations with language a promising and easily accessible way to improve the generalization of vision models in low-data settings.

\section*{Acknowledgments}
We thank Pang Wei Koh, Sebastian Schuster, and Dan Iter for helpful discussions and feedback, Mike Wu and Jacob Andreas for discussions and code, and our anonymous reviewers for insightful comments.
This work was supported by an NSF Graduate Research Fellowship for JM, a SAIL-Toyota Research Award, and the Office of Naval Research grant ONR MURI N00014-16-1-2007.
Toyota Research Institute (TRI) provided funds to assist the authors with their research but this article solely reflects the opinions and conclusions of its authors and not TRI or any other Toyota entity.

\section*{Reproducibility}
Code, data, and experiments are available at \url{https://github.com/jayelm/lsl} and on CodaLab at \url{https://bit.ly/lsl_acl20}.

\bibliography{langrepr_acl20}

\begin{thebibliography}{32}
\expandafter\ifx\csname natexlab\endcsname\relax\def\natexlab#1{#1}\fi

\bibitem[{Andreas et~al.(2018)Andreas, Klein, and Levine}]{andreas2018learning}
Jacob Andreas, Dan Klein, and Sergey Levine. 2018.
\newblock Learning with latent language.
\newblock In \emph{Proceedings of the 2018 Conference of the North American
  Chapter of the Association for Computational Linguistics: Human Language
  Technologies, Volume 1 (Long Papers)}, pages 2166--2179.

\bibitem[{Bahdanau et~al.(2019)Bahdanau, Hill, Leike, Hughes, Hosseini, Kohli,
  and Grefenstette}]{bahdanau2019learning}
Dzmitry Bahdanau, Felix Hill, Jan Leike, Edward Hughes, Arian Hosseini,
  Pushmeet Kohli, and Edward Grefenstette. 2019.
\newblock Learning to understand goal specifications by modelling reward.
\newblock In \emph{International Conference on Learning Representations
  (ICLR)}.

\bibitem[{Camburu et~al.(2018)Camburu, Rockt{\"a}schel, Lukasiewicz, and
  Blunsom}]{camburu2018snli}
Oana-Maria Camburu, Tim Rockt{\"a}schel, Thomas Lukasiewicz, and Phil Blunsom.
  2018.
\newblock e-{SNLI}: natural language inference with natural language
  explanations.
\newblock In \emph{Advances in Neural Information Processing Systems
  (NeurIPS)}, pages 9539--9549.

\bibitem[{Chen et~al.(2019)Chen, Liu, Kira, Wang, and Huang}]{chen2019closer}
Wei-Yu Chen, Yen-Cheng Liu, Zsolt Kira, Yu-Chiang~Frank Wang, and Jia-Bin
  Huang. 2019.
\newblock A closer look at few-shot classification.
\newblock In \emph{International Conference on Learning Representations
  (ICLR)}.

\bibitem[{Cho et~al.(2014)Cho, van Merrienboer, Gulcehre, Bahdanau, Bougares,
  Schwenk, and Bengio}]{cho2014learning}
Kyunghyun Cho, Bart van Merrienboer, Caglar Gulcehre, Dzmitry Bahdanau, Fethi
  Bougares, Holger Schwenk, and Yoshua Bengio. 2014.
\newblock Learning phrase representations using {RNN} encoder--decoder for
  statistical machine translation.
\newblock In \emph{Proceedings of the 2014 Conference on Empirical Methods in
  Natural Language Processing (EMNLP)}, pages 1724--1734.

\bibitem[{Chopra et~al.(2019)Chopra, Tessler, and Goodman}]{chopra2019ratchet}
Sahil Chopra, Michael~Henry Tessler, and Noah~D Goodman. 2019.
\newblock The first crank of the cultural ratchet: Learning and transmitting
  concepts through language.
\newblock In \emph{Proceedings of the 41st Annual Meeting of the Cognitive
  Science Society}, pages 226--232.

\bibitem[{Copestake et~al.(2016)Copestake, Emerson, Goodman, Horvat, Kuhnle,
  and Muszynska}]{copestake2016resources}
Ann~A Copestake, Guy Emerson, Michael~Wayne Goodman, Matic Horvat, Alexander
  Kuhnle, and Ewa Muszynska. 2016.
\newblock Resources for building applications with dependency minimal recursion
  semantics.
\newblock In \emph{International Conference on Language Resources and
  Evaluation (LREC)}.

\bibitem[{Donahue and Grauman(2011)}]{donahue2011annotator}
Jeff Donahue and Kristen Grauman. 2011.
\newblock Annotator rationales for visual recognition.
\newblock In \emph{Proceedings of the IEEE International Conference on Computer
  Vision (ICCV)}, pages 1395--1402.

\bibitem[{Frome et~al.(2013)Frome, Corrado, Shlens, Bengio, Dean, Marc'Aurelio,
  and Mikolov}]{frome2013devise}
Andrea Frome, Greg~S Corrado, Jon Shlens, Samy Bengio, Jeff Dean, Ranzato
  Marc'Aurelio, and Tomas Mikolov. 2013.
\newblock {DeViSE}: A deep visual-semantic embedding model.
\newblock In \emph{Advances in Neural Information Processing Systems
  (NeurIPS)}, pages 2121--2129.

\bibitem[{Goodman(1955)}]{goodman1955fact}
Nelson Goodman. 1955.
\newblock \emph{Fact, fiction, and forecast}.
\newblock Harvard University Press, Cambridge, MA.

\bibitem[{Gordo and Larlus(2017)}]{gordo2017beyond}
Albert Gordo and Diane Larlus. 2017.
\newblock Beyond instance-level image retrieval: Leveraging captions to learn a
  global visual representation for semantic retrieval.
\newblock In \emph{Proceedings of the IEEE Conference on Computer Vision and
  Pattern Recognition (CVPR)}, pages 6589--6598.

\bibitem[{Goyal et~al.(2019)Goyal, Niekum, and Mooney}]{goyal2019using}
Prasoon Goyal, Scott Niekum, and Raymond~J. Mooney. 2019.
\newblock Using natural language for reward shaping in reinforcement learning.
\newblock In \emph{Proceedings of the Twenty-Eighth International Joint
  Conference on Artificial Intelligence, {IJCAI-19}}, pages 2385--2391.

\bibitem[{Hancock et~al.(2018)Hancock, Varma, Wang, Bringmann, Liang, and
  R{\'e}}]{hancock2018training}
Braden Hancock, Paroma Varma, Stephanie Wang, Martin Bringmann, Percy Liang,
  and Christopher R{\'e}. 2018.
\newblock Training classifiers with natural language explanations.
\newblock In \emph{Proceedings of the 56th Annual Meeting of the Association
  for Computational Linguistics (Volume 1: Long Papers)}, pages 1884--1895.

\bibitem[{He et~al.(2016)He, Zhang, Ren, and Sun}]{he2016deep}
Kaiming He, Xiangyu Zhang, Shaoqing Ren, and Jian Sun. 2016.
\newblock Deep residual learning for image recognition.
\newblock In \emph{Proceedings of the IEEE Conference on Computer Vision and
  Pattern Recognition (CVPR)}, pages 770--778.

\bibitem[{He and Peng(2017)}]{he2017fine}
Xiangteng He and Yuxin Peng. 2017.
\newblock Fine-grained image classification via combining vision and language.
\newblock In \emph{Proceedings of the IEEE Conference on Computer Vision and
  Pattern Recognition (CVPR)}, pages 5994--6002.

\bibitem[{Hendricks et~al.(2016)Hendricks, Akata, Rohrbach, Donahue, Schiele,
  and Darrell}]{hendricks2016generating}
Lisa~Anne Hendricks, Zeynep Akata, Marcus Rohrbach, Jeff Donahue, Bernt
  Schiele, and Trevor Darrell. 2016.
\newblock Generating visual explanations.
\newblock In \emph{Proceedings of the European Conference on Computer Vision
  (ECCV)}, pages 3--19.

\bibitem[{Hendricks et~al.(2018)Hendricks, Hu, Darrell, and
  Akata}]{hendricks2018grounding}
Lisa~Anne Hendricks, Ronghang Hu, Trevor Darrell, and Zeynep Akata. 2018.
\newblock Grounding visual explanations.
\newblock In \emph{Proceedings of the European Conference on Computer Vision
  (ECCV)}, pages 264--279.

\bibitem[{Kingma and Ba(2015)}]{kingma2014adam}
Diederik~P Kingma and Jimmy Ba. 2015.
\newblock Adam: A method for stochastic optimization.
\newblock In \emph{International Conference on Learning Representations
  (ICLR)}.

\bibitem[{Kuhnle and Copestake(2017)}]{kuhnle2017shapeworld}
Alexander Kuhnle and Ann Copestake. 2017.
\newblock Shapeworld-a new test methodology for multimodal language
  understanding.
\newblock \emph{arXiv preprint arXiv:1704.04517}.

\bibitem[{Pennington et~al.(2014)Pennington, Socher, and
  Manning}]{pennington2014glove}
Jeffrey Pennington, Richard Socher, and Christopher Manning. 2014.
\newblock Glo{V}e: Global vectors for word representation.
\newblock In \emph{Proceedings of the 2014 Conference on Empirical Methods in
  Natural Language Processing (EMNLP)}, pages 1532--1543.

\bibitem[{Rajani et~al.(2019)Rajani, McCann, Xiong, and
  Socher}]{rajani2019explain}
Nazneen~Fatema Rajani, Bryan McCann, Caiming Xiong, and Richard Socher. 2019.
\newblock Explain yourself! {L}everaging language models for commonsense
  reasoning.
\newblock In \emph{Proceedings of the 57th Annual Meeting of the Association
  for Computational Linguistics (ACL)}, pages 4932--4942, Florence, Italy.

\bibitem[{Reed et~al.(2016)Reed, Akata, Lee, and Schiele}]{reed2016learning}
Scott Reed, Zeynep Akata, Honglak Lee, and Bernt Schiele. 2016.
\newblock Learning deep representations of fine-grained visual descriptions.
\newblock In \emph{Proceedings of the IEEE Conference on Computer Vision and
  Pattern Recognition (CVPR)}, pages 49--58.

\bibitem[{Simonyan and Zisserman(2015)}]{simonyan2014very}
Karen Simonyan and Andrew Zisserman. 2015.
\newblock Very deep convolutional networks for large-scale image recognition.
\newblock In \emph{International Conference on Learning Representations
  (ICLR)}.

\bibitem[{Snell et~al.(2017)Snell, Swersky, and Zemel}]{snell2017prototypical}
Jake Snell, Kevin Swersky, and Richard Zemel. 2017.
\newblock Prototypical networks for few-shot learning.
\newblock In \emph{Advances in Neural Information Processing Systems
  (NeurIPS)}, pages 4077--4087.

\bibitem[{Socher et~al.(2013)Socher, Ganjoo, Manning, and Ng}]{socher2013zero}
Richard Socher, Milind Ganjoo, Christopher~D Manning, and Andrew Ng. 2013.
\newblock Zero-shot learning through cross-modal transfer.
\newblock In \emph{Advances in Neural Information Processing Systems
  (NeurIPS)}, pages 935--943.

\bibitem[{Srivastava et~al.(2017)Srivastava, Labutov, and
  Mitchell}]{srivastava2017joint}
Shashank Srivastava, Igor Labutov, and Tom Mitchell. 2017.
\newblock Joint concept learning and semantic parsing from natural language
  explanations.
\newblock In \emph{Proceedings of the 2017 Conference on Empirical Methods in
  Natural Language Processing (EMNLP)}, pages 1527--1536.

\bibitem[{Tokmakov et~al.(2019)Tokmakov, Wang, and
  Hebert}]{tokmakov2019learning}
Pavel Tokmakov, Yu-Xiong Wang, and Martial Hebert. 2019.
\newblock Learning compositional representations for few-shot recognition.
\newblock In \emph{Proceedings of the IEEE International Conference on Computer
  Vision (ICCV)}, pages 6372--6381.

\bibitem[{Tomasello(1999)}]{tomasello1999cultural}
Michael Tomasello. 1999.
\newblock \emph{The Cultural Origins of Human Cognition}.
\newblock Harvard University Press, Cambridge, MA.

\bibitem[{Vapnik and Vashist(2009)}]{vapnik2009new}
Vladimir Vapnik and Akshay Vashist. 2009.
\newblock A new learning paradigm: Learning using privileged information.
\newblock \emph{Neural Networks}, 22(5-6):544--557.

\bibitem[{Wah et~al.(2011)Wah, Branson, Welinder, Perona, and
  Belongie}]{wah2011caltech}
Catherine Wah, Steve Branson, Peter Welinder, Pietro Perona, and Serge
  Belongie. 2011.
\newblock The {Caltech-UCSD} {B}irds-200-2011 dataset.

\bibitem[{Xing et~al.(2019)Xing, Rostamzadeh, Oreshkin, and
  Pinheiro}]{xing2019adaptive}
Chen Xing, Negar Rostamzadeh, Boris Oreshkin, and Pedro~O Pinheiro. 2019.
\newblock Adaptive cross-modal few-shot learning.
\newblock In \emph{Advances in Neural Information Processing Systems
  (NeurIPS)}, pages 4848--4858.

\bibitem[{Zaidan et~al.(2007)Zaidan, Eisner, and Piatko}]{zaidan2007using}
Omar Zaidan, Jason Eisner, and Christine Piatko. 2007.
\newblock Using “annotator rationales” to improve machine learning for text
  categorization.
\newblock In \emph{Human Language Technologies 2007: The Conference of the
  North {A}merican Chapter of the Association for Computational Linguistics;
  Proceedings of the Main Conference (NAACL-HLT)}, pages 260--267.

\end{thebibliography}
\bibliographystyle{acl_natbib}

\onecolumn
\newpage
\appendix
\twocolumn

\section{Model and training details}
\label{app:details}

\subsection{ShapeWorld}

\paragraph{$f_\theta$.} Like \citet{andreas2018learning}, $f_\theta$ starts with features extracted from the last convolutional layer of a fixed ImageNet-pretrained VGG-19 network \cite{simonyan2014very}. These 4608-d embeddings are then fed into two fully connected layers $\in \mathbb{R}^{4608 \times 512}, \mathbb{R}^{512 \times 512}$ with one ReLU nonlinearity in between.

\paragraph{LSL.} For LSL, the 512-d embedding from $f_\theta$ directly initializes the 512-d hidden state of the GRU $g_\phi$. We use 300-d word embeddings initialized randomly. Initializing with GloVe \cite{pennington2014glove} made no significant difference.

\paragraph{L3.} $f_\theta$ and $g_\phi$ are the same as in LSL and Meta. $h_\eta$ is a unidirectional 1-layer GRU with hidden size 512 sharing the same word embeddings as $g_\phi$. The output of the last hidden state is taken as the embedding of the description $\mathbf{w}^{(t)}$. Like \citet{andreas2018learning}, a total of 10 descriptions per task are sampled at test time.

\paragraph{Training.} We train for 50 epochs, each epoch consisting of 100 batches with 100 tasks in each batch, with the Adam optimizer \cite{kingma2014adam} and a learning rate of $0.001$. We select the model with highest epoch validation accuracy during training. This differs slightly from \citet{andreas2018learning}, who use different numbers of epochs per model and did not specify how they were chosen; otherwise, the training and evaluation process is the same.

\paragraph{Data.} We recreated the ShapeWorld dataset using the same code as \citet{andreas2018learning}, except generating 4x as many test tasks (4000 vs 1000) for more stable confidence intervals.

Note that results for both L3 \emph{and the baseline model} (Meta) are 3--4 points lower than the scores reported in \citet{andreas2018learning} (because performance is lower for all models, we are not being unfair to L3). This is likely due to differences in model initialization due to our PyTorch reimplementation and/or recreation of the dataset with more test tasks.

\subsection{Birds}
\label{app:birds_details}

\paragraph{$f_\theta$.} The 4-layer convolutional backbone $f_\theta$ is the same as the one used in much of the few-shot literature \citep{chen2019closer,snell2017prototypical}. The model has 4 convolutional blocks, each consisting of a 64-filter 3x3 convolution, batch normalization, ReLU nonlinearity, and 2x2 max-pooling layer. With an input image size of $84 \times 84$ this results in 1600-d image embeddings. Finally, the bilinear matrix $\mathbf{W}$ used in the similarity function has dimension $1600 \times 1600$. 

\paragraph{LSL.} The resulting 1600-d image embeddings are fed into a single linear layer $\in \mathbb{R}^{1600 \times 200}$ which initializes the 200-d hidden state of the GRU. We initialize embeddings with GloVe. We did not observe significant gains from increasing the size of the decoder $g_\phi$.

\paragraph{L3.} $f_\theta$ and $g_\phi$ are the same. $h_\eta$ is a unidirectional GRU with hidden size 200 sharing the same embeddings as $g_\phi$. The last hidden state is taken as the concept $\mathbf{c}_n$. 10 descriptions per class are sampled at test time. We did not observe significant gains from increasing the size of the decoder $g_\phi$ or encoder $h_\eta$, nor increasing the number of descriptions sampled per class at test.

\paragraph{Training.} For ease of comparison to the few-shot literature we use the same training and evaluation process as \citet{chen2019closer}. Models are trained for 60000 episodes, each episode consisting of one randomly sampled task with 16 query images per class. Like \citet{chen2019closer}, they are evaluated on 600 episodes. We use Adam with a learning rate of 0.001 and select the model with the highest validation accuracy after training.

\textbf{Data.} Like \citet{chen2019closer}, we use standard data preprocessing and training augmentation: ImageNet mean pixel normalization, random cropping, horizontal flipping, and color jittering.

\end{document}